\documentclass[10pt,twocolumn,letterpaper]{article}

\usepackage{cvpr}              

\usepackage{graphicx}
\usepackage{float} 
\usepackage{caption}
\usepackage{amsmath}
\usepackage{amssymb}
\usepackage{booktabs}
\usepackage{enumitem}
\usepackage{xcolor, color, colortbl}

\usepackage{url}

\usepackage{graphicx}
\usepackage{mathtools}
\usepackage{amssymb}
\usepackage{amsthm}
\usepackage{booktabs}
\usepackage{bbm}
\usepackage{wrapfig}
\usepackage{amsmath, bm}
\usepackage{algorithm, algpseudocode}
\usepackage{algorithmicx}
\usepackage{comment}

\newcommand{\cy}[1]{{\color{orange}{cy: #1}}}

\newcommand{\eps}{{\epsilon}}

\newcommand\NAME{\texttt{DpDy}\xspace}

\usepackage[pagebackref,breaklinks,colorlinks]{hyperref}
\usepackage{bm}

\definecolor{Highlight}{HTML}{39b54a}  

\usepackage{amssymb}
\usepackage{pifont}
\usepackage{algorithm}
\usepackage{listings}
\usepackage{etoolbox}
\makeatletter
\AfterEndEnvironment{algorithm}{\let\@algcomment\relax}
\AtEndEnvironment{algorithm}{\kern2pt\hrule\relax\vskip3pt\@algcomment}
\let\@algcomment\relax
\newcommand\algcomment[1]{\def\@algcomment{\footnotesize#1}}
\renewcommand\fs@ruled{\def\@fs@cfont{\bfseries}\let\@fs@capt\floatc@ruled
  \def\@fs@pre{\hrule height.8pt depth0pt \kern2pt}%
  \def\@fs@post{}%
  \def\@fs@mid{\kern2pt\hrule\kern2pt}%
  \let\@fs@iftopcapt\iftrue}
\makeatother
\newcommand{\cmmnt}[1]{}

\usepackage[normalem]{ulem}

\usepackage[capitalize]{cleveref}
\crefname{section}{Sec.}{Secs.}
\Crefname{section}{Section}{Sections}
\Crefname{table}{Table}{Tables}
\crefname{table}{Tab.}{Tabs.}

\def\cvprPaperID{3805} 
\def\confName{CVPR}
\def\confYear{2024}

\begin{document}


\title{%
  Diffusion Priors for Dynamic View Synthesis from Monocular Videos
}



\author{
Chaoyang Wang$^{1}$
\hspace*{1em}
Peiye Zhuang$^{1}$
\hspace*{1em}
Aliaksandr Siarohin$^{1}$
\hspace*{1em}
Junli Cao$^{1}$
\\
\hspace*{1em}
Guocheng Qian$^{1,2}$
\hspace*{1em}
Hsin-Ying Lee$^{1}$
\hspace*{1em}
Sergey Tulyakov$^{1
}$
\\
$^{1}$Snap Research
\hspace*{1em}
$^{2}$KAUST
\\
\small 
\{\texttt{cwang9, pzhuang, asiarohin, jcao2, gqian, hlee5, stulyakov}\}\texttt{@snap.com}
}


\twocolumn[{
\maketitle
\begin{center}
\includegraphics[width=1.0\textwidth]{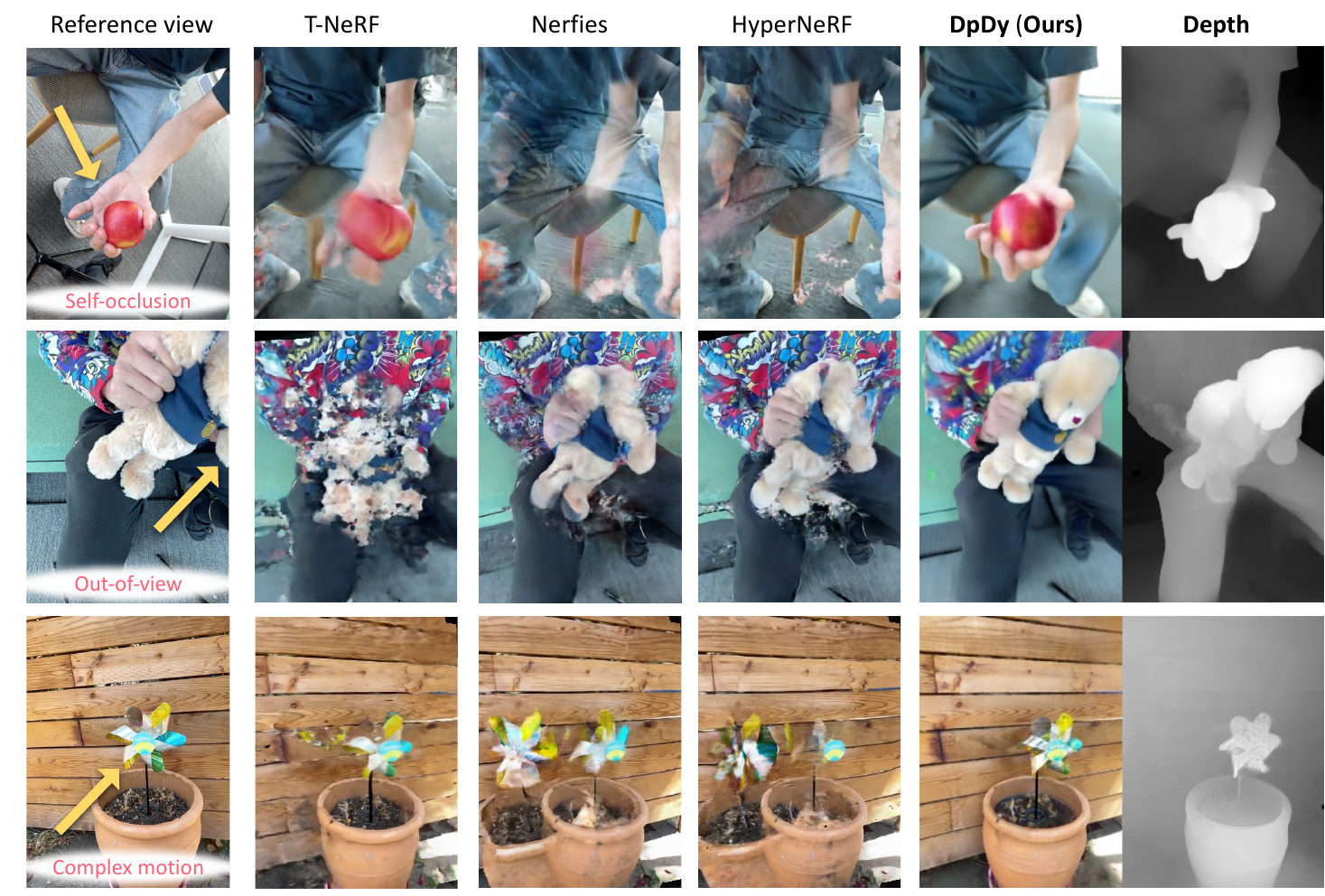}
\captionof{figure}
{\textbf{Comparison with dynamic novel view synthesis methods from monocular videos}. When dealing with self-occlusions, out-of-view details, and complex motions previous methods render severe artifacts (columns 2-4). In contrast, our novel approach based on diffusion prior elegantly handles such cases, producing high quality visual results and geometry (right two columns).
}
\label{fig:teaser}
\end{center}
}]

\makeatletter

\makeatother



\begin{abstract}
Dynamic novel view synthesis aims to capture the temporal evolution of visual content within videos. 
Existing methods struggle to distinguishing between motion and structure, particularly in scenarios where camera poses are either unknown or constrained compared to object motion.
Furthermore, with information solely from reference images, it is extremely challenging to hallucinate unseen regions that are occluded or partially observed in the given videos.
To address these issues, we first finetune a pretrained RGB-D diffusion model on the video frames using a customization technique. 
Subsequently, we distill the knowledge from the finetuned model to a 4D representations encompassing both dynamic and static Neural Radiance Fields (NeRF) components. 
The proposed pipeline achieves geometric consistency while preserving the scene identity. 
We perform thorough experiments to evaluate the efficacy of the proposed method qualitatively and quantitatively. Our results demonstrate the robustness and utility of our approach in challenging cases, further advancing dynamic novel view synthesis. 
Our project website is at \url{https://mightychaos.github.io/dpdy_proj/}.

\end{abstract}

\section{Introduction}

The novel view synthesis of dynamic scenes from monocular casual videos holds significant importance in various domains due to its potential impact on understanding and interacting with the real world.
While existing methods approach this challenge through the utilization of hand-crafted geometric and physics priors~\cite{Nerfies, HyperNeRF} or by leveraging monocular depth estimation~\cite{nsff, liu2023robust}, recent analyses~\cite{dycheck} underscore the limitations of both paradigms. 
Hand-crafted geometric and physics priors  prove insufficient in disambiguating motion and structure,  particularly in in-the-wild scenarios where the camera motion is smaller than the object motion, and methods relying on monocular depth estimation tends to produce paper-thin foreground objects and do not provide effective supervision for occluded regions, leading to severe artifacts when the dynamic object is close to the camera.

A critical challenge in dynamic novel view synthesis is to hallucinate regions unseen in videos, where existing methods struggle when relying solely on information from reference views.
There common scenarios contribute to this challenge.
First, regions behind visible surfaces in the reference views cannot be recovered in novel views.
Second, some parts of objects are entirely out of view in reference images.
Third, without sufficient information from enough camera poses, some objects cannot be realistically reconstructed.
We demonstrate these challenges in Fig.~\ref{fig:teaser}.

To address the need for information beyond reference images in the given video, leveraging prior knowledge from by pretrained models emerges as a potential solution.
Recently advancements in 3D reconstruction from a single image, facing similar challenges, have witnessed a great progress by distilling knowledge from large-scale 2D text-to-image diffusion models as a 2D prior to help synthesize unseen regions~\cite{clipmesh, DreamFusion, magic3d, SJC, fantasia3d, HIFA, ProlificDreamer}.
More recently, the 3D consistency of the reconstructed objects has been further improved with the help of multi-view diffusion models~\cite{tang2023mvdiffusion,Zero-1-to-3}, finetuned on 3D object data.
Despite sharing similar challenges, these techniques are not directly applicable to dynamic novel view synthesis.
First, the multi-view models are trained on object-centric and static data, and cannot handle scenes that are complex and dynamic. 
Second, a domain gap exists between the training images of these diffusion models and the real-world in-the-wild images, hindering direct knowledge distillation while maintaining consistency.

In response to these challenges, we propose \NAME, an effective dynamic novel view synthesis pipeline leveraging geometry priors from pretrained diffusion models with customization techniques.
First, we represent a 4D scene with a dynamic NeRF for dynamic motions and a rigid NeRF for static regions. 
To achieve \textbf{geometry consistency}, we integrate knowledge distillation~\cite{DreamFusion,HIFA} from a pretrained RGB-D image diffusion model~\cite{ldm3d} in addition to the conventional reconstruction objective.
Moreover, to preserve the scene \textbf{identity} and to mitigate the domain gap, we finetune the RGB-D diffusion model using video frames with customization techniques~\cite{dreambooth}. 

We conduct extensive qualitative and quantitative experiments on the challenging iPhone dataset~\cite{gao2022monocular}, featuring diverse and complex motions. 
We evaluate the quality of the 4D reconstruction using masked Learned Perceptual Image Patch Similarity (LPIPS)~\cite{LPIPS} and masked Structural Similarity Index (SSIM) scores.
\NAME performs favorably against all baseline methods. However, we found that the standard metrics do not adequately reflect the quality of the rendered novel views. Hence, we performed a series of user studies against previous works. The human annotators almost unanimously selected our method in almost all comparisons, supporting the benefits of using 2D diffusion priors for dynamic novel view supervision. 


\section{Related Works}

\noindent\textbf{Dynamic View Synthesis from Monocular Videos} involves learning a 4D representation from a casual monocular video. Previous works~\cite{nsff, HyperNeRF, liu2023robust, xian2021space,wang2021neural,Gao_2021_ICCV,TiNeuVox,Cao2022FWD,Yang_2023_CVPR} typically employ a \textit{dynamic} Neural Radiance Field (D-NeRF)~\cite{NeRF} as a 4D representation that encodes spatio-temporal scene contents. These approaches use hand-crafted geometric and physics priors to learn the 4D representations.
For instance, flow-based methods like NSFF~\cite{nsff,liu2023robust} utilize a scene flow field warping loss to enforce temporal consistency. Canonical-based methods~\cite{Nerfies, HyperNeRF, pumarola2021d,Wang_2023_CVPR, lei2022cadex}, represent a scene using a deformation field that maps each local observation to a canonical scene space. 
Most of these methods are limited to object-centric scenes with controlled camera poses. More recently, DyniBaR~\cite{li2023dynibar} extends the multi-view conditioned NeRF approach, i.e., IBRNet~\cite{wang2021ibrnet}, to allow dynamic novel view synthesis with a freely chosen input camera trajectory. Another practical obstacle for applying dynamic NeRFs to real-world videos is the robust estimation of camera poses when videos contain a large portion of moving objects. Recent works, such as RoDynRF~\cite{liu2023robust}, propose a joint optimization of dynamic NeRF and camera poses, demonstrating practicality in real-world applications.

Despite showing promising results, we note that hand-crafted priors such as deformation and temporal smoothness, utilized by prior works, are insufficient for reconstructing complex 4D scenes from monocular videos. This is due to the complexity of resolving ambiguities between motion and structure, as well as hallucinating unseen or occluded regions. This challenge is particularly pronounced with slow-motion cameras or in scenarios involving complex dynamic motions, such as those found in the DyCheck dataset~\cite{dycheck}. To overcome these limitations, we integrate large-scale image diffusion priors to effectively hallucinate the unseen regions within a scene and regularize 4D reconstruction.

\vspace{.1cm}

\begin{figure*}
\centering
 \includegraphics[width=\linewidth]{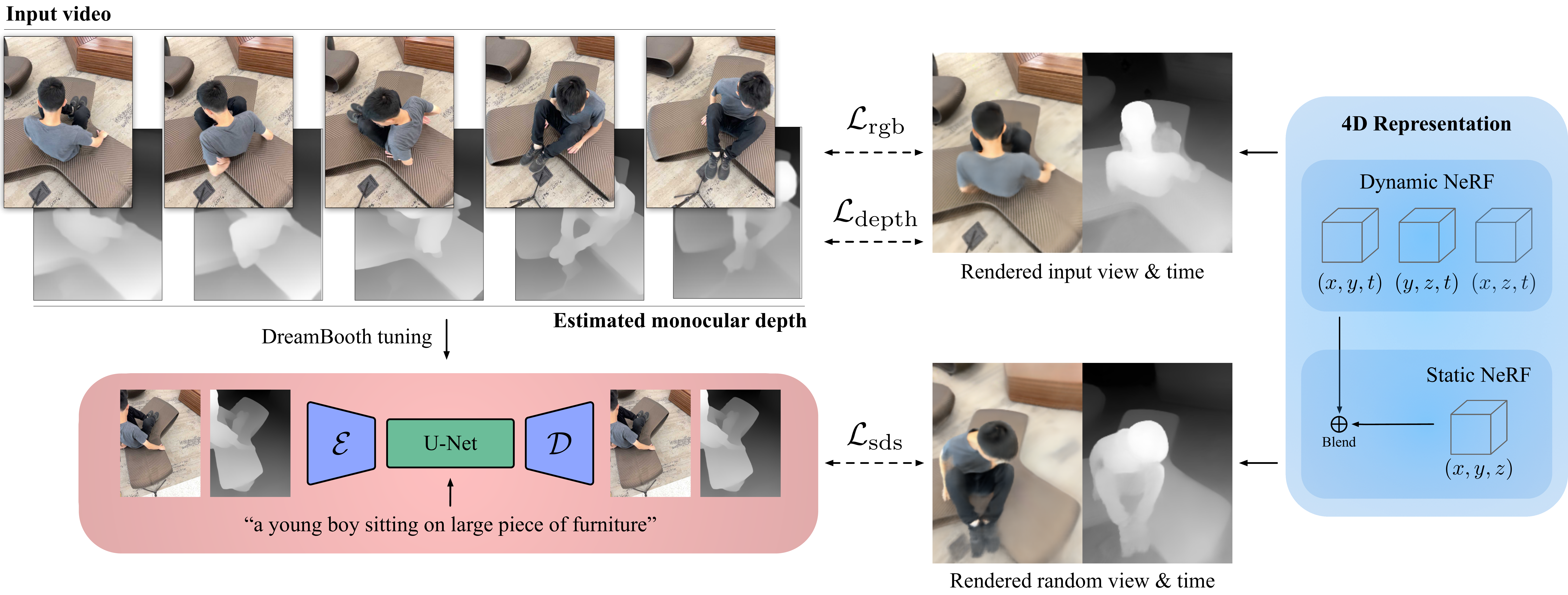}
 \vspace{-20pt}
 \caption{\textbf{Overview of our method.}
 To perform dynamic novel view synthesis given a video, we adopt a 4D representation consisting of dynamic and static parts. We use two types of supervision. First, we render the input viewpoints at input time. Besides, we distill prior knowledge of a pre-trained RGB-D diffusion model on random novel views using score distillation sampling.
 Furthermore, to mitigate the domain gaps between the training distributions and in-the-wild images, we tune the RGB-D diffusion model using the reference images with a customization technique prior to distillation. }
 \label{fig:method}
\end{figure*}

\noindent\textbf{Text-to-Image Diffusion Priors} refer to large-scale text-to-image diffusion-based generative models~\cite{imagen, stablediffusion}. These models provide large-scale 2D image priors that can benefit 3D and 4D generation tasks which struggle with data limitation issues. For instance, recent text-to-3D generation works~\cite{clipmesh, jain2021dreamfields, DreamFusion, makeit3d, magic3d, SJC, fantasia3d, HIFA, ProlificDreamer,scenetex} have successfully achieved high-quality 3D asset generation by using image guidance from 2D image priors to the 3D domain. To achieve this, a Score Distillation Sampling (SDS) approach~\cite{DreamFusion} is introduced, where noise is added to an image rendered from the 3D representation and subsequently denoised by a pre-trained text-to-image generative model. SDS minimizes the Kullback–Leibler (KL) divergence between a prior Gaussian noise distribution and the estimated noise distribution.
Additionally, image-to-3D generation works~\cite{Magic123,Zero-1-to-3,RealFusion,long2023wonder3d} also use text-to-image diffusion priors. Differently, these works have additional requirements where the image identity should be kept. For this purpose, Dreambooth~\cite{dreambooth} is proposed to fine-tune the UNet and text encoders in a text-to-image diffusion model based on the given image. Dreambooth fine-tuning enables the diffusion prior to memorize the given image identity, thus providing meaningful guidance for image-based 3D reconstruction.

While these 3D generation approaches based on text-to-image diffusion priors have experienced rapid development, their focus has primarily been on generating 3D assets rather than real-world scenes or videos. Inspired by this, we extend their application to the in-the-wild 4D scene reconstruction task and incorporate a pre-trained RGB-D diffusion model, LDM3D~\cite{ldm3d}, as an RGB-D prior to hallucinate unseen views of 4D scenes.

\section{Method}

We aim to achieve 4D dynamic novel view synthesis from monocular videos. To achieve this, we propose our method as illustrated in Fig.~\ref{fig:method}. 
Specifically, we represent a 4D dynamic scene using two separate NeRFs~\cite{mildenhall2020nerf}: one for rigid regions and another for dynamic regions of a scene. The rendering of images involves blending the output from these two NeRF fields.
To optimize the NeRF representations, we apply reconstruction losses for images and depth maps to minimize the difference between rendered images and the reference video frames (Sec.~\ref{sec:recon}). Additionally, we use an SDS loss in the joint RGB and depth (a.k.a. RGB-D) space to provide guidance for novel dynamic view synthesis (Sec.~\ref{sec:prior}). Formally, we define the loss function $\mathcal{L}$ as:

\begin{equation}
\mathcal{L} = \lambda_\text{rgb} \mathcal{L}_\text{rgb} + \lambda_\text{depth} \mathcal{L}_\text{depth} + \lambda_\text{reg} \mathcal{L}_\text{reg} +  \lambda_\text{sds} \mathcal{L}_\text{sds} .
\end{equation}

Here, $\mathcal{L}_\text{rgb}$ denotes the image-space reconstruction loss on seen video frames. $\mathcal{L}_\text{depth}$ represents an affine-invariant depth reconstruction loss using a pre-trained depth prediction model~\cite{ranftl2020towards}. 
Additionally, we incorporate a regularization loss $\mathcal{L}_\text{reg}$ to regularize the 4D representation. 
Finally, $\mathcal{L}_\text{sds}$ is an SDS loss for novel dynamic views in RGB-D space.
We present our technical details in the following. 


\subsection{4D Representation}
\label{sec:recon}

We represent a 4D scene with two separate NeRFs: the \textit{static} NeRF focuses on representing the static regions and the the \textit{dynamic} NeRF aims to capture the dynamic motions  of the scene. Formally, $\bm c_s, \sigma_s, \bm c_d, \sigma_d$ denotes the color and density of a point on a ray from the static and the dynamic NeRF, respectively. Our method is not specific to the exact implementation of NeRFs. The details of our implementation is provided in Sec.~\ref{sec:impl_details}.

Given a camera pose and a timestep, we obtain an image from the NeRFs via differentiable rendering. The color of an image pixel along a NeRF ray $r$, denoted as $C(r)$, is volumetrically rendered from the blended radiance field of the static and dynamic NeRFs. The rendering equation is formally written as 
\begin{equation}
    C(r) = \int_{t_n}^{t_f} T(t) \left[ \sigma_s(t) \bm c_s(t) + \sigma_d(t) \bm c_d(t) \right] dt,
    \label{eq:2}
\end{equation}
where $T(t) = \exp(-\int_{t_n}^t (\sigma_s(s) + \sigma_d(s))ds)$ is the accumulated transmittance, and  $t_n$ and $t_f$ are the near and far bounds of a ray. We derived the following numerical estimator for the continuous integral in Eq.~\ref{eq:2}:
\begin{equation}
\begin{aligned}
    &C(r) = \sum_{i=1}^N T_i (1-\exp(-(\sigma_{s_i}+\sigma_{d_i})\delta_i)\bm c_i, \\
    &\text{where} \quad T_i = \exp \left( - \sum_{j=1}^{i-1} (\sigma_{s_j} + \sigma_{d_j}) \delta_j \right), \\
    & \bm c_i = \frac{\sigma_{s_i} \bm c_{s_i} + \sigma_{d_i} \bm  c_{d_i}}{\sigma_{s_i} + \sigma_{d_i}} \quad
    \text{and}\quad \delta_i = t_{i+1} - t_i.
    \label{eq:3}
\end{aligned}
\end{equation}
This discretized rendering equation differs from prior discretization approaches~\cite{nerfinthewild, nsff} that need separate accumulations of the static and dynamic components. Eq.~\ref{eq:3} is computationally more efficient as it can be implemented by calling upon the standard NeRF volumetric rendering just once. Please refer to the supplementary material for mathematical proof.



\noindent \textbf{Reconstruction Losses.} We render images and the corresponding depth map from the NeRFs. Subsequently, we calculate an image reconstruction loss $\mathcal{L}_\text{rgb}$ using the $L_1$ norm and an affine-invariant depth loss $\mathcal{L}_\text{depth}$ by comparing the rendered depth map with a pre-computed depth map obtained from the off-the-shelf monocular depth estimator MiDAS~\cite{MiDaS}. It is worth noting that the depth estimation from MiDAS is both noisy and non-metric, and lacking temporal consistency. As a result, we only incorporate it during the initial training process and progressively reduce the weight of the depth loss $\mathcal{L}_\text{depth}$ over training. We mention that the two reconstruction losses are only applied on existing views that can be seen from the videos. In Sec.~\ref{sec:prior}, we introduce depth guidance for unseen views.


\noindent \textbf{Regularization.}
We apply additional regularization to the 4D NeRF representation. The regularization loss $\mathcal{L}_\text{reg}$ consists of two parts which we present in Eq.~\ref{eq:4}-~\ref{eq:5}, respectively. \emph{First}, to promote concentration of radiance near the visible surface, we employ the z-variance loss~\cite{HIFA}, penalizing the weighted sum of square distances from points on the ray to the mean depth, \ie,
\begin{equation}
    \sum_i (z_i - \mu_z)^2 \frac{w_i}{\sum_j w_j}, \quad \text{where} \quad \mu_z = \sum_i \frac{w_i z_i}{\sum_j w_j},
    \label{eq:4}
\end{equation}
where $z_i$ is the depth for each sampled points along the ray, $\mu_z$ is rendered depth and $w_i$ is the normalized volumetric rendering weight. 

\emph{Second}, to encourage proper decomposition of dynamic foreground and static background, we penalize the skewed entropy of the foreground-background density ratio $\frac{\sigma_d}{\sigma_d + \sigma_s}$, as proposed by D$^2$-NeRF~\cite{D2nerf}. Specifically, the loss is written as:
\begin{equation}
    H((\frac{\sigma_d}{\sigma_d + \sigma_s})^k), 
\label{eq:5}
\end{equation}
where $H(x) = -( x \log(x) + (1-x) \log(1 -x) )$ is a binary entropy loss. The skew parameter $k$ is set to 2, promoting separation biased towards increasing background regions.

\begin{table*}[t]
\caption{
\textbf{Novel view synthesis results.}
We compare the mLPIPS and mSSIM scores with existing methods on the iPhone dataset~\cite{gao2022monocular}.
}
\vspace{-2mm}
\label{tab:quantitative_iphone}
\centering
\resizebox{1\linewidth}{!} 
{
\begin{tabular}{l ccccccc|c}
\toprule
%
 mLPIPS $\downarrow$ / mSSIM $\uparrow$  & Apple & Block & Paper-windmill & Space-out & Spin & Teddy & Wheel & Macro-average \\
\midrule
%
T-NeRF + Lidar~\cite{gao2022monocular} & 0.508 / 0.728 & 0.346 / 0.669 & 0.258 / 0.367 &  0.377 / 0.591 & 0.443 / 0.567 & 0.429 / 0.570 & 0.292 / 0.548 & 0.379 / 0.577 \\
NSFF + Lidar~\cite{li2021neural} & 0.478 / 0.750 & 0.389 / 0.639 & 0.211 / 0.378 & 0.303 / 0.622 & 0.309 / 0.585 & 0.372 / 0.557 & 0.310 / 0.458 & 0.339 / 0.569 \\
\midrule
T-NeRF~\cite{gao2022monocular} &0.581 / 0.712 &0.441/ 0.629 & 0.444 / 0.302 &0.408 / 0.593 & 0.491 / 0.508 & 0.472 / 0.555 & 0.441 / 0.629 & 0.468 / 0.561\\
Nerfies~\cite{Nerfies} & 0.610 / 0.703 & 0.550 / 0.569  & 0.506 / 0.277 & 0.440 / 0.546 &0.385 / 0.533 &0.460 / 0.542 & 0.535 / 0.326 & 0.498 / 0.500\\
HyperNeRF~\cite{HyperNeRF} & 0.601 / 0.696 & 0.517 / 0.586 &0.501 / 0.272 &  0.437 / 0.554 &0.547 / 0.444 & 0.397 / 0.556 & 0.547 / 0.322 &0.507 / 0.490 \\
RoDynRF~\cite{liu2023robust} & 0.552 / 0.722 & 0.513 / 0.634 & 0.482 / 0.321 & 0.413 / 0.594 & 0.570 / 0.484 & 0.613 / 0.536 & 0.478 / 0.449 & 0.517 / 0.534 \\
\NAME (Ours) & 0.596 / \textbf{0.735} & 0.478 / 0.630 &  0.447 / \textbf{0.387} & 0.457 / \textbf{0.622} & 0.571 / 0.500 & 0.562 / 0.531 & 0.504 / 0.511 & 0.516 / 0.559 \\
\bottomrule
\end{tabular}
}

\end{table*}

\subsection{Diffusion Priors for Novel View Supervision}
\label{sec:prior}

As aforementioned, using reconstruction losses on existing views is insufficient. To address this challenge, we employ guidance from RGB-D diffusion priors for novel views of the 4D scenes. Using RGB-D diffusion priors offers two advantages. Firstly, comparing to the use of RGB diffusion priors in previous text-to-3D generation work~\cite{DreamFusion}, RGB-D guidance provides direct geometry supervision. 
Moreover, unlike depth guidance using off-the-shelf monocular depth estimation, which produces a \textit{single} certain depth map conditioned on a given image, the RGB-D diffusion model learns a joint distribution of images and depth maps. 
As a result, the RGB-D diffusion model provides as output a \textit{distribution} of image-depth pairs, enabling more robust supervision for 4D scene reconstruction.

Practically, we use LDM3D~\cite{ldm3d} as the RGB-D prior. An LDM3D model is a \textit{latent} diffusion model that consists of an encoder $\mathcal{E}$, a decoder $\mathcal{D}$, and a denoising function $\epsilon_\phi$.
The encoder $\mathcal{E}$ compresses the input RGB-D image $\bm x$ into a low-resolution latent vector $\bm z$, denoted as $\bm z \coloneqq \mathcal{E}(\bm x)$, and the decoder $\mathcal{D}$ recovers the RGB-D image from the latent vector $\bm z$.
The denoising score function $\epsilon_\phi$ predicts the given noise on the latent vector that has been perturbed by noise $\bm \eps$, the estimated noise denoted as $\bm{\hat{\eps}}$. 
Formally, we denote the gradient of the SDS loss $\mathcal{L}_\text{sds}$ as:

\begin{equation}
\begin{aligned}
\nabla_\theta \mathcal{L}_\text{sds} = \mathbb{E}_{t, \bm \eps}~\omega(t) (\bm{\hat{\eps}} - \bm \eps) \frac{\partial \bm z}{\partial \bm x} \frac{\partial \bm x}{\partial \theta} ,
\end{aligned}
\end{equation}
where $\theta$ represents for the parameters of the NeRFs and $\omega(t)$ is a weighting function. 

Note that the input depth map of the LDM3D model is up-to-affine. Thus, we normalize the NeRF-rendered depth maps to $0-1$ range.

\noindent \textbf{Personalization.} Similar to recent image-to-3D generation work~\cite{Magic123}, we apply the Dreambooth fine-tuning approach~\cite{dreambooth} to refine the LDM3D model using the given monocular videos. 
Specifically, we fine-tune the UNet diffusion model and the text encoder in the LDM3D model.
The text prompt is automatically generated by using BLIP~\cite{BLIP-2}.
To obtain the depth map of the video frames, we use a pre-trained depth estimation model, MiDaS~\cite{MiDaS}. 
Since the output depth from MiDaS is affine-invairant, we apply a $0-1$ normalization on the predicted depth maps from MiDaS before the fine-tuning process.

\begin{figure}[ht!]
    \centering
    \includegraphics[width=\linewidth]{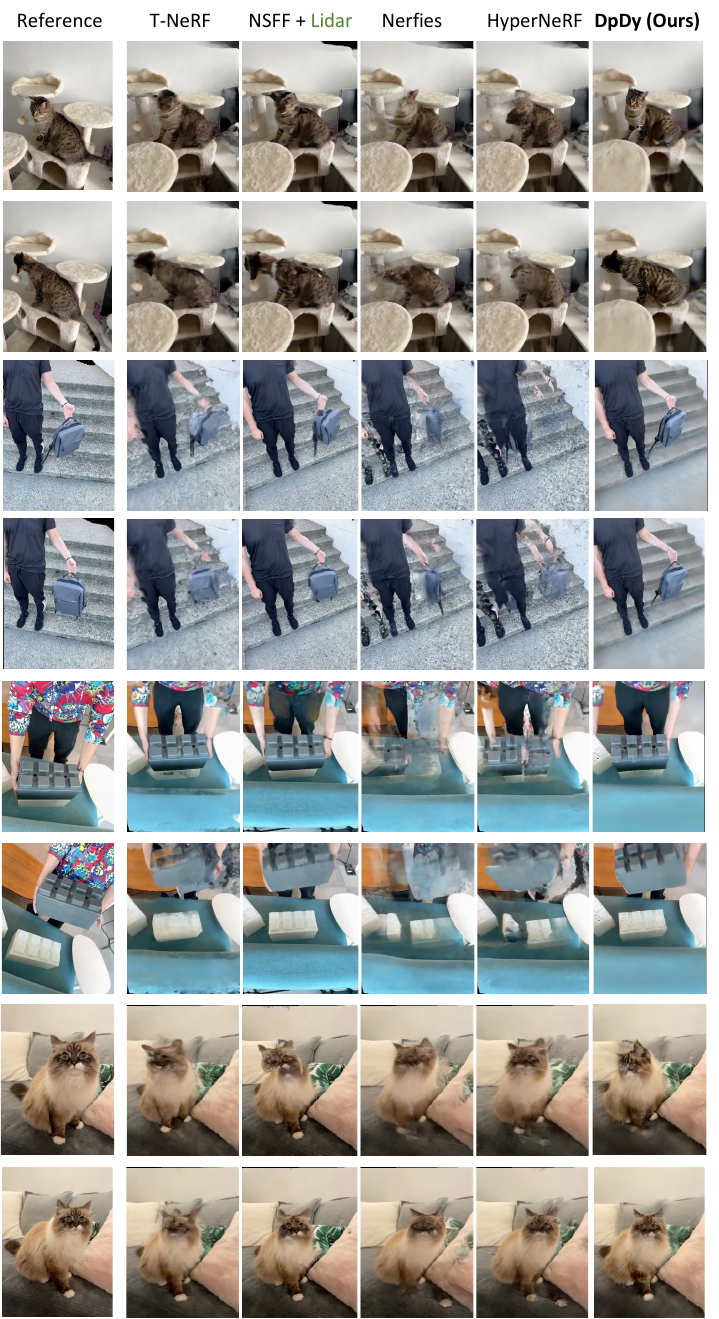}
    \caption{\textbf{Qualitative comparison on the iPhone dataset.} For each sequence, we sample two frames to show case view synthesis result under different object motions. The left-most column displays reference image frames used during training, while the images on the right showcase rendering results for a novel viewpoint. Our method excels in producing the most realistic view synthesis for dynamic foregrounds, surpassing the baseline method (NSFF~\cite{nsff}) that incorporates Lidar depth as extra supervision. It is important to note that while our background maintains geometric consistency, it appears blurrier compared to the baselines. This observation explains why our method does not show an advantage when evaluated using image-based metrics such as SSIM and LPIPS. However, in a user study (see Table~\ref{tab:user-study}) focusing on video quality inspection, our method significantly outperforms the baselines. }
    \label{fig:comparison1}
\end{figure}

\section{Experiments}
\subsection{Implementation Details}
\label{sec:impl_details}
\noindent\textbf{Dynamic NeRF Representation} The static and dynamic component of our 4D NeRF representation is built on multi-level hash grids \ie instant-NGP~\cite{InstantNGP}. The static component is a standard hash grid. For the dynamic component, we chose to decompose 4D space-time into three subspaces. Specifically, we have three hash grids, each encodes xyz, xyt, yzt, xzt subspaces. The resulting encodings from these hash grids are concatenated and then passed through small MLPs to produce output colors and densities. The decomposition of 4D into lower-dimensional subspaces has been explored in previous works~\cite{Cao2022FWD,kplanes_2023}. We observe that different implementations of such decomposition do not significantly impact final results. Therefore, we choose the implementation with the lowest rendering time. 

During training, we render $240\times140$-res image, 1/4th of the original image size.
To improve rendering efficiency, we employ the importance sampling with a proposal density network as in MipNeRF 360~\cite{MipNeRF-360}. The small proposal density network (modeled using hash grids as described above, but with smaller resolution and cheaper MLPs) samples 128 points per ray and the final NeRF model samples 64 points. 
Additional detailed hyperparameters of our 4D representation are provided in the supplementary.

\noindent\textbf{Optimization Details.} The selection of hyperparameters, specifically $\lambda_\text{rgb}$ and $\lambda_\text{sds}$, plays a crucial role in shaping the training dynamics. A high value for $\lambda_\text{rgb}$ tends to result in slow improvement on novel views. Conversely, a large $\lambda_\text{sds}$ can lead to the loss of video identity during the initial stages of training. To strike a balance,  we choose to initiate training with a substantial $\lambda_\text{rgb}$ value set to 1.0, emphasizing the accurate fitting of input video frames in the early phase. Subsequently, we decrease it to 0.1, shifting the focus towards enhancing novel views in the later stages of training. $\lambda_\text{sds}$ is kept fixed as 1.0 throughout all training iterations.

We empirically set the weighting for z-variance as 0.1 and the skewed entropy of the foreground-background density ratio as 1e-3. The method is trained for 30,000 iterations with a fixed learning rate of 0.01 using the Adam optimizer.

\noindent\textbf{SDS Details.} We adopt the noise scheduling approach introduced in HIFA~\cite{HIFA}. Specifically, instead of uniformly sample $t$ as in standard SDS~\cite{DreamFusion}, we find deterministically anneal $t$ from 0.6 to 0.2 leads to finer reconstruction details. We set a small classifier-free guidance weight as 7.5. Using larger weights, such as 100, would result in over-saturated result.

\noindent\textbf{Virtual View Sampling.} During training iterations, we randomly sample a camera viewpoint for novel view supervision by perturbing a camera position randomly picked from input video frames. 360$^\circ$ viewpoint sampling is currently not supported due to the complexity of real-world scenes, making it challenging to automatically sample cameras that avoid occlusion while keeping the object of interest centered. A more principled approach is deferred to future work.

\subsection{Evaluation}
\label{sec:evaluation}
\noindent\textbf{Dataset}. We conduct experiments on the iPhone dataset from DyCheck~\cite{dycheck}. It contains 14 videos, among which half of them comes with held-out camera views for quantitative evaluation. This dataset presents a more challenging and realistic scenario compared to widely-used datasets such as Nvidia Dynamic View Synthesis~\cite{Nvidia-dataset} and the data proposed by Nerfies~\cite{Nerfies} and HyperNeRF~\cite{HyperNeRF}. The iPhone dataset features natural camera motion, in contrast to other datasets where cameras teleport between positions. Gao~\etal~\cite{dycheck} discovered that methods performing well in teleported video sequences experience a significant performance drop on the iPhone dataset. Teleported videos makes the target dynamic scene appears quasi-static, thus makes the problem simpler, but less practical since everyday video captured by users usually does not contain rapid camera motions.

\noindent\textbf{Baselines.} We compare against well-known methods including NSFF~\cite{nsff}, Nerfies~\cite{Nerfies}, HyperNeRF~\cite{HyperNeRF}, T-NeRF~\cite{gao2022monocular}, and more recent approach \ie RoDynRF~\cite{liu2023robust}. Baselines reported in the DyCheck paper were improved through supervision with a Lidar depth sensor (denoted with ``+ Lidar"). 
Given our method's commitment to practicality without assuming the use of depth sensors, our primary focus lies in comparing against baselines that \emph{do not} involve Lidar depth. We also made a sincere attempt to reproduce DynIBaR~\cite{li2023dynibar}. However, due to the complexity of their implementation, which includes undisclosed third-party modules, we were unable to generate reasonable results on the iPhone dataset. As a result, it was omitted from our comparison.

\noindent\textbf{Metrics.} We adopt the evaluation metrics proposed by Gao~\etal~\cite{dycheck}, including masked Learned Perceptual Image Patch Similarity (mLPIPS)~\cite{LPIPS} and Structural Similarity Index (mSSIM) scores, focusing on regions co-visible between testing and training views. However, we find that these metrics do not reflect the perceived quality of novel views. For instance, the baseline method involving the training of a time-modulated NeRF (T-NeRF) without advanced regularization attains the best performance according to the metrics. However, a visual inspection reveals that T-NeRF produces least satisfactory results in dynamic regions, often resulting in fragmented or blurry foregrounds. This discrepency between visual quality and testing metrics is due to methods using only input monocular videos is thereotically not possible to estimate the correct relative scale between moving foreground and static background. The scale ambiguity introduces shifts, enlargements, or shrinkages in the rendered foreground compared to ground-truth images, leading to decreases in SSIM and LPIPS metrics. However, it does not notably impact the perceived quality for human. Creating a metric that better aligns with perceived visual quality is a non-trivial task, and we leave this for future research.


\subsection{Comparison to Baseline Methods}

\begin{figure}
    \centering
    \includegraphics[width=\linewidth]{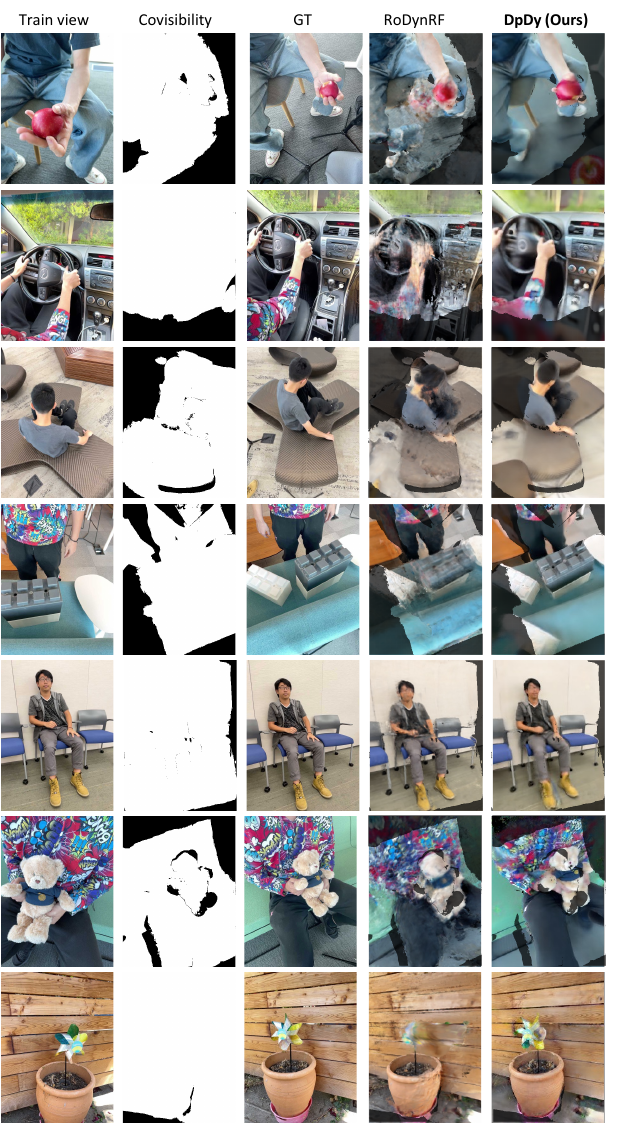}
    \caption{\textbf{Qualitive comparison against RoDynRF~\cite{liu2023robust}.} We visualize results for all testing sequences in the iPhone dataset following the evaluation protocol. We masked out regions outside the provided covisibility mask. Our method demonstrates greater realism and fewer artifacts compared to RoDynRF.}
    \label{fig:comparison2}
\end{figure}

We present qualitative comparison results in Fig.~\ref{fig:comparison1}-\ref{fig:comparison2}, where we maintain a fixed camera pose identical to the first video frame and render the subsequent frames.  We observe that our method produces the most visually satisfying results among all the compared methods. Canonical-based methods, such as Nerfies and HyperNeRF, exhibit limited flexibility in capturing complex or rapid object motions, such as finger interactions with an apple or the motion of a circulating paper windmill. T-NeRF consistently produces noisy results during rapid object motions, as seen when a person quickly shifts a teddy bear. The baseline with the closest visual quality to ours is NSFF supervised with Lidar depth. While its background is relatively more stable, its foreground is often blurrier and more flickering compared to ours. Please refer to our supplementary video for more detailed visual comparison.

We quantitatively compare our method with the baseline methods, as shown in Table~\ref{tab:quantitative_iphone}.
The baselines are categorized into two groups based on whether Lidar depth is used. Row 2-3 showcase results using NSFF~\cite{nsff} and T-NeRF~\cite{gao2022monocular} with Lidar depth incorporated during training.
Rows 4-8 present results where Lidar depth is not employed. We find our method is competitive in terms mLPIPS and mSSIM. However, as previously discussed in Sec.~\ref{sec:evaluation}, we note the metrics do not reflect the perceived visual quality. As depicted in Fig.\ref{fig:comparison2}, RoDyRF~\cite{liu2023robust} yields similar metrics to ours. However, their rendering under test views exhibits numerous artifacts and severe flickering when inspecting the rendered videos. We attribute the reason why our method does not exhibit a significant advantage over RoDyRF in terms of metrics to our results being slightly over-smoothed and having color drift due to the current limitations of SDS -- which is shared with other SDS-based text-to-3D generation methods. Additionally, as discussed in Sec.\ref{sec:evaluation}, dynamic regions are slightly shifted (most noticeable in the human example in the 3rd row of Fig.~\ref{fig:comparison2}) compared to the ground truth due to scale ambiguities, rendering metrics unable to fully capture the visual quality.

\begin{table}[t]
\caption{
\textbf{User study results.}
We report the percentage of annotators choosing our method against a competing baseline. Two different settings are reported: bullet-time and stabilized-view rendering.
}
\vspace{-8pt}
\label{tab:user-study}
\centering
\resizebox{1\linewidth}{!} 
{
\begin{tabular}{l ccc}
\toprule
Experiment & T-NeRF~\cite{gao2022monocular} & Nerfies~\cite{Nerfies} & HyperNeRF~\cite{HyperNeRF} \\
\midrule
Bullet-time rendering & 97\% & 100\% & 97\% \\
Stabilized-view rendering & 97\% & 100\% & 83\% \\

\bottomrule
\end{tabular}
}

\end{table}

\subsection{User Studies}

Commonly adopted metric do not precisely reflect the advantages of our method as discussed before. To compare methods visually, we performed a user study, in which the annotators were asked to select a visualization that is most consistent volumetrically, has least ghost artifacts, and overall looks more realistic. Table~\ref{tab:user-study} reports the results. We performed two different experiments. In the first, we rendered the dynamic scene using bullet-time effect. In the second, we stabilized the view. Our method is preferred by human annotators in almost all the cases. This clearly shows the advantages of the proposed approach.

\subsection{Ablation Studies}

\begin{figure}
    \centering
    \includegraphics[width=\linewidth]{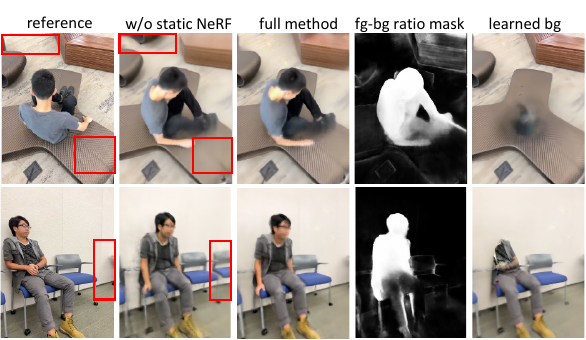}
    \caption{\textbf{Ablation on static-dynamic component decomposition.} Our method incorporates a dynamic foreground and static background decomposition module, a critical element for synthesizing sharper backgrounds with reduced hallucination, as highlighted by the red rectangles. }
    \label{fig:abl-fgbg}
\end{figure}

\noindent\textbf{Ablation on Static-Dynamic Component Decomposition.} 
We find that decomposing the static and dynamic components in our 4D representation is crucial for achieving satisfactory visual results. Without the static component, the background exhibits flickering and is more susceptible to SDS hallucinating non-existent objects on the dynamically changing background, as highlighted in Fig.~\ref{fig:abl-fgbg} using red boxes. In contrast, our full method achieves a clean separation between static and dynamic elements, resulting in a more stable background with fewer hallucinated contents.

\begin{figure}
    \centering
    \includegraphics[width=\linewidth]{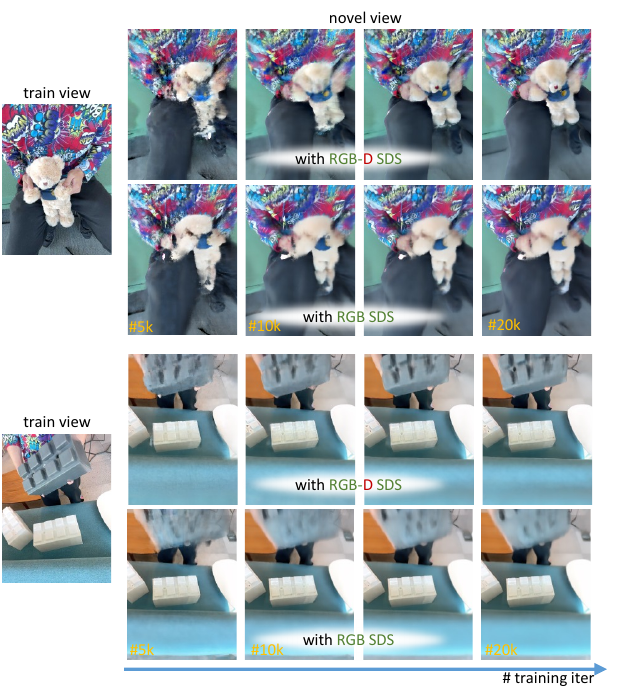}
    \caption{\textbf{Ablation on RGB-D v.s. RGB Diffusion Priors.} We visualize the rendering of novel views at every 5,000 training iterations. Using RGB-D SDS produces superior results compared to using RGB SDS.}
    \label{fig:abl-sds}
\end{figure}

\noindent\textbf{Ablation on RGB-D Diffusion Priors.} We present results by using RGB diffusion priors, instead of the RGB-D prior. 
Specifically, the RGB prior is obtained from a pre-trained RealisticVision V5.1 model\footnote{https://civitai.com/models/4201/realistic-vision-v20} which has been shown to leads to more realistic text-3D generation~\cite{HIFA} compared to StableDiffusion 1.5~\cite{stablediffusion}. 
We keep the same hyperparameters of DreamBooth for both RGB and RGB-D model finetuning. 
Fig.~\ref{fig:abl-sds} shows the comparison. Compared to using the RGB prior (2nd row of each example), the novel view results with the RGB-D prior (1st row of each example) exhibit more detailed texture, reasonable geometries, and fast convergence over training iterations.

\section{Conclusion}

We propose a novel approach for dynamic 4D scene reconstruction from monocular videos. Unlike previous works that encounter challenges when employing hand-crafted priors for generating novel views, to the best of our knowledge, our method is the first to explore 2D image diffusion priors for dynamic novel view supervision in generic scenes. This incorporation enhances the robustness of our method in addressing challenges such as self-occlusion, out-of-view occlusion, and complex object motions. Our findings suggest that future research should leverage the advantages brought by large generative models.

Despite encouraging results, our method has \textbf{limitations} summarized as follows:
\textbf{(1)} Due to the necessity of rendering the entire image and running large diffusion models, our method currently requires high-end GPUs for over 10 hours of training per video with 400 frames. Constrained by computational cost, the resolution of our view synthesis is limited. Future works could explore more efficient representations, such as Gaussian splatting~\cite{wu20234d, kerbl20233d}, and lighter diffusion models~\cite{li2023snapfusion, kim2023bk};
\textbf{(2)} Temporal smoothness is currently implicitly regularized by the multi-level design of instant-NGP. It may not be robust enough for flickering-free reconstruction. Although we have preliminarily explored utilizing video diffusion priors (e.g., AnimateDiff~\cite{animatediff}) in SDS loss, substantial improvement was not observed. We leave the exploration of stronger video diffusion models as future work;
\textbf{(3)} The current implementation is confined to a bounded dynamic scene. Extending this work to an unbounded scene can be achieved through either progressively combining multiple grids~\cite{wang2023f2} or using image-conditioned rendering, as in DyniBaR~\cite{li2023dynibar}.
\textbf{(4)} Finetuning on single video losses generalization for diffusion models. Currently, our method does not support 360$^\circ$ reconstruction if the input video did not already enumerate the surrounding views. Image-conditioned generative models could potentially eliminate the need for finetuning, but currently available models are trained on object-centric data with backgrounds removed. 

\section*{Supplementary Material}
In the supplementary material, additional implementation details are provided. For more visual results, please refer to the webpage at \url{https://mightychaos.github.io/dpdy_proj/}.

\section*{A: 4D Representation Details}
\begin{figure}[h]
    \centering
    \includegraphics[width=\linewidth]{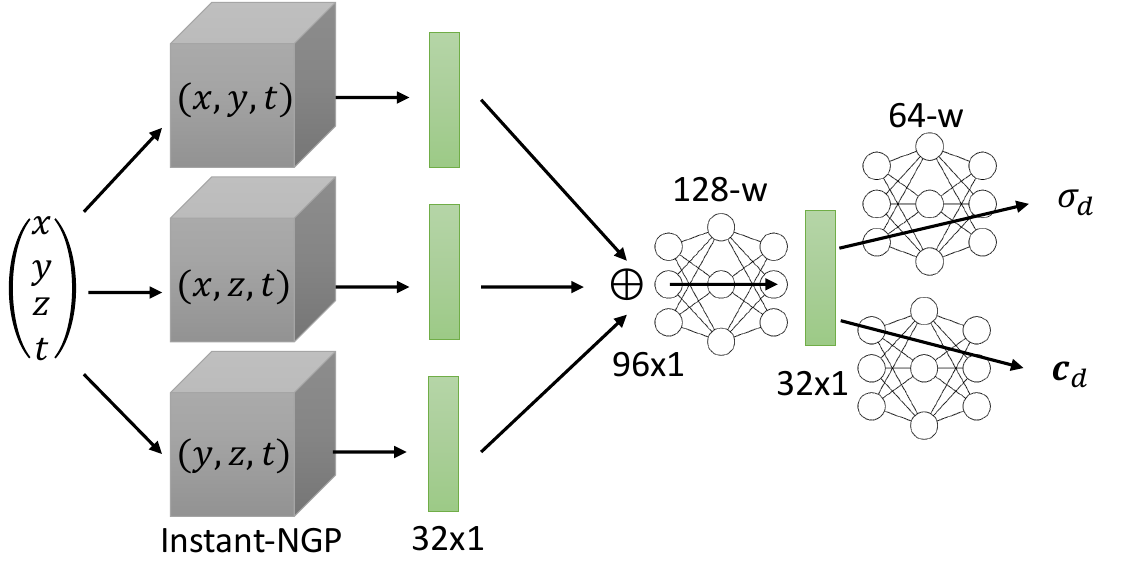}
    \caption{Illustration of using 3D grids to represent 4D spacetime radiance fields. }
    \label{fig:dyn_nerf_illu}
\end{figure}

\noindent\textbf{4D representation with 3D grids.}
Motivated by recent works (\cite{kplanes_2023,Cao2022FWD,Chen2022ECCV}) that decompose high-dimensional voxels into lower-dimensional planes, in this work we choose to decompose the 4D space-time grid into three 3D grids. Each 3D grid is represented using an instant-NGP, capturing the $(x, y, t)$, $(x, z, t)$, and $(y, z, t)$ subspaces. The hyperparameters of the instant-NGPs are detailed in Table \ref{tab:hash_grid_param}. To extract the density and color information of a spacetime point $(x, y, z, t)$, depicted in Fig.~\ref{fig:dyn_nerf_illu}, we query each of the three 3D grids, obtaining three embeddings. Subsequently, these embeddings are concatenated and input into a small MLP to yield a fused embedding. This fused representation is then directed through additional MLPs to generate predictions for density ($\sigma_d$) and color ($\mathbf{c}_d$).

\begin{table}[h]
    \caption{Hyper-parameter of Instant-NGP for 4D representation.}
    \centering
    \begin{tabular}{c|c|c}
    \toprule
         & Density Proposal & Radiance Field \\
    \midrule
    n\_levels  & 8 & 16\\
    n\_features\_per\_level   & 2 & 2\\
    log2\_hashmap\_size & 19 & 19\\
    base\_resolution & 16 & 16\\
    per\_level\_scale & 1.447 & 1.447\\
    \bottomrule
    \end{tabular}
    \label{tab:hash_grid_param}
\end{table}

\noindent\textbf{Blending radiance fields.}
The color of an image pixel along a ray $r$, denoted as $C(r)$, is rendered from the blended radiance field of the static and dynamic NeRFs with densities $\sigma_s$, $\sigma_d$ and colors $\mathbf{c}_s$, $\mathbf{c}_d$.
\begin{equation}
    C(r) = \int_{t_n}^{t_f} T(t) \left[ \sigma_s(t) \bm c_s(t) + \sigma_d(t) \bm c_d(t) \right] dt,
    \label{eq:2}
\end{equation}
where $T(t) = \exp(-\int_{t_n}^t (\sigma_s(s) + \sigma_d(s))ds)$ is the accumulated transmittance, and  $t_n$ and $t_f$ are the near and far bounds of a ray. 

The discretized equation of Eq.~\ref{eq:2} is computed as follow:
\begin{equation}
\begin{aligned}
    &C(r) = \sum_{i=1}^N T_i (1-\exp(-(\sigma_{s_i}+\sigma_{d_i})\delta_i)\bm c_i, \\
    &\text{where} \quad T_i = \exp \left( - \sum_{j=1}^{i-1} (\sigma_{s_j} + \sigma_{d_j}) \delta_j \right), \\
    & \bm c_i = \frac{\sigma_{s_i} \bm c_{s_i} + \sigma_{d_i} \bm  c_{d_i}}{\sigma_{s_i} + \sigma_{d_i}} \quad
    \text{and}\quad \delta_i = t_{i+1} - t_i.
    \label{eq:3}
\end{aligned}
\end{equation}

\begin{proof}
\begin{equation*}
    \begin{aligned}
        C(r) = & \int_{t_n}^{t_f} T(t) \left[ \sigma_s(t) \bm c_s(t) + \sigma_d(t) \bm c_d(t) \right] dt \\
        = & \int_{t_n}^{t_f} e^{-\int_{t_n}^t (\sigma_s(s) + \sigma_d(s))ds} \left[ \sigma_s(t) \bm c_s(t) + \sigma_d(t) \bm c_d(t) \right] dt \\
        = & \int_{t_n}^{t_f} \frac{d}{dt}e^{-\int_{t_n}^t (\sigma_s(s) + \sigma_d(s))ds} \frac{\sigma_s(t) \bm c_s(t) + \sigma_d(t) \bm c_d(t)}{\sigma_s(t) + \sigma_d(t)} dt \\
        \approx & \sum_{i=1}^N (T_i - T_{i+1}) \frac{\sigma_{s_i} \bm c_{s_i} + \sigma_{d_i} \bm  c_{d_i}}{\sigma_{s_i} + \sigma_{d_i}} \\
        = & \sum_{i=1}^N T_i (1 - \exp(-(\sigma_{si}+\sigma_{di})\delta_i))\frac{\sigma_{s_i} \bm c_{s_i} + \sigma_{d_i} \bm  c_{d_i}}{\sigma_{s_i} + \sigma_{d_i}} 
    \end{aligned}
\end{equation*}
\end{proof}

In practise, to avoid numerical instability, we add a small value $\epsilon=1e-6$ to the denominator of $\mathbf{c}_i$, \ie,
\begin{equation}
\bm c_i = \frac{\sigma_{s_i} \bm c_{s_i} + \sigma_{d_i} \bm  c_{d_i}}{\sigma_{s_i} + \sigma_{d_i} + \epsilon}.
\end{equation}

\section*{B: Training Details}
\noindent \textbf{Rendering details.}
For each training iteration, we render one reference view image and one novel view image. Since diffusion models are finetuned using $512\times512$-res images, we randomly crop the novel view image into a square image patch and resize it to $512\times512$-res. To match the distribution of the pretrained RGB-D diffusion model, we convert the rendered depth map to disparity map by taking the reciprocal of depth values. Then the disparity map is normalized between 0 and 1 by:
\begin{equation}
    d = \frac{d-d_\text{min}}{d_\text{max}-d_\text{min} + \epsilon},
\end{equation}
where $d_\text{max}$ and $d_\text{min}$ are $95\%$-percentile values, which are more robust to noise compared to directly taking the maximum and minimum disparity values.

\noindent \textbf{Hyperparameters.} We employ the same set of hyperparameters across all experiments, with detailed weightings for each loss function provided in Table~\ref{tab:hyperparameter}.

\begin{table}[h]
    \centering
    \begin{tabular}{c|cc}
    \toprule
        param. & value & decay step\\
    \midrule
        $\lambda_\text{rgb}$ & 1.0 $\Longrightarrow$ 0.1 & 7k\\
        $\lambda_\text{depth}$ & 0.1 $\Longrightarrow$ 0.01 & 2k \\   
        $\lambda_\text{z-variance}$ & 0.1 & -\\ 
        $\lambda_\text{f-g decomp}$ & 1e-4 & - \\ 
        $\lambda_\text{sds}$ & 1.0 & -\\ 
    \bottomrule
    \end{tabular}
    \caption{Hyperparameter of the loss function. $\Longrightarrow$ denotes the weighting of the loss exponentially decays. }
    \label{tab:hyperparameter}
\end{table}

{\small
\bibliographystyle{ieee_fullname}
\bibliography{reference.bbl}
}

\end{document}